\documentclass[runningheads]{llncs}

\usepackage{eccv}

\usepackage{eccvabbrv}

\usepackage{graphicx}
\usepackage{booktabs}
\usepackage{xcolor}
\definecolor{lightgreen}{rgb}{0.76, 0.93, 0.56}
\usepackage{transparent}
\usepackage{colortbl}
\usepackage{pifont}

\usepackage[accsupp]{axessibility}  

\usepackage[pagebackref,breaklinks,colorlinks,citecolor=eccvblue]{hyperref}

\usepackage{orcidlink}
\usepackage{multirow}
\usepackage{multicol}
\usepackage{textcomp}
\begin{document}

\title{ELLA: Equip Diffusion Models with LLM for Enhanced Semantic Alignment} 

\titlerunning{ELLA}


\author{Xiwei Hu\thanks{Equal Contribution} \and
Rui Wang\inst{\star} \and
Yixiao Fang\inst{\star}\and
Bin Fu\inst{\star}\and
Pei Cheng\and
Gang Yu\thanks{Corresponding Author}
}

\authorrunning{X. Hu, R. Wang, Y. Fang, B. Fu, P. Cheng, and G. Yu}



\institute{Tencent\\
\email{xiweihu@outlook.com, \{raywwang, yixiaofang, brianfu, peicheng\}@tencent.com, skicy@outlook.com}\\
\url{https://ella-diffusion.github.io}
}
\maketitle

\begin{abstract}
  Diffusion models have demonstrated remarkable performance in the domain of text-to-image generation. However, most widely used models still employ CLIP as their text encoder, which constrains their ability to comprehend dense prompts, encompassing multiple objects, detailed attributes, complex relationships, long-text alignment{\it, etc}.
  In this paper, we introduce an \textbf{E}fficient \textbf{L}arge \textbf{L}anguage Model \textbf{A}dapter, termed \textbf{ELLA}, which equips text-to-image diffusion models with powerful Large Language Models (LLM) to enhance text alignment {\it without training of either U-Net or LLM}. 
  To seamlessly bridge two pre-trained models, we investigate a range of semantic alignment connector designs and propose a novel module, the Timestep-Aware Semantic Connector (TSC), which dynamically extracts timestep-dependent conditions from LLM.
  Our approach adapts semantic features at different stages of the denoising process, assisting diffusion models in interpreting lengthy and intricate prompts over sampling timesteps.
  Additionally, ELLA can be readily incorporated with community models and  
  tools to improve their prompt-following capabilities.
  To assess text-to-image models in dense prompt following, we introduce Dense Prompt Graph Benchmark (DPG-Bench), a challenging benchmark consisting of 1K dense prompts. 
  Extensive experiments demonstrate the superiority of ELLA in dense prompt following compared to state-of-the-art methods,
  particularly in multiple object compositions involving diverse attributes and relationships.

\end{abstract}

\begin{figure}[!t]
  \centering
  \begin{subfigure}{1\linewidth}
    \includegraphics[width=0.99\linewidth]{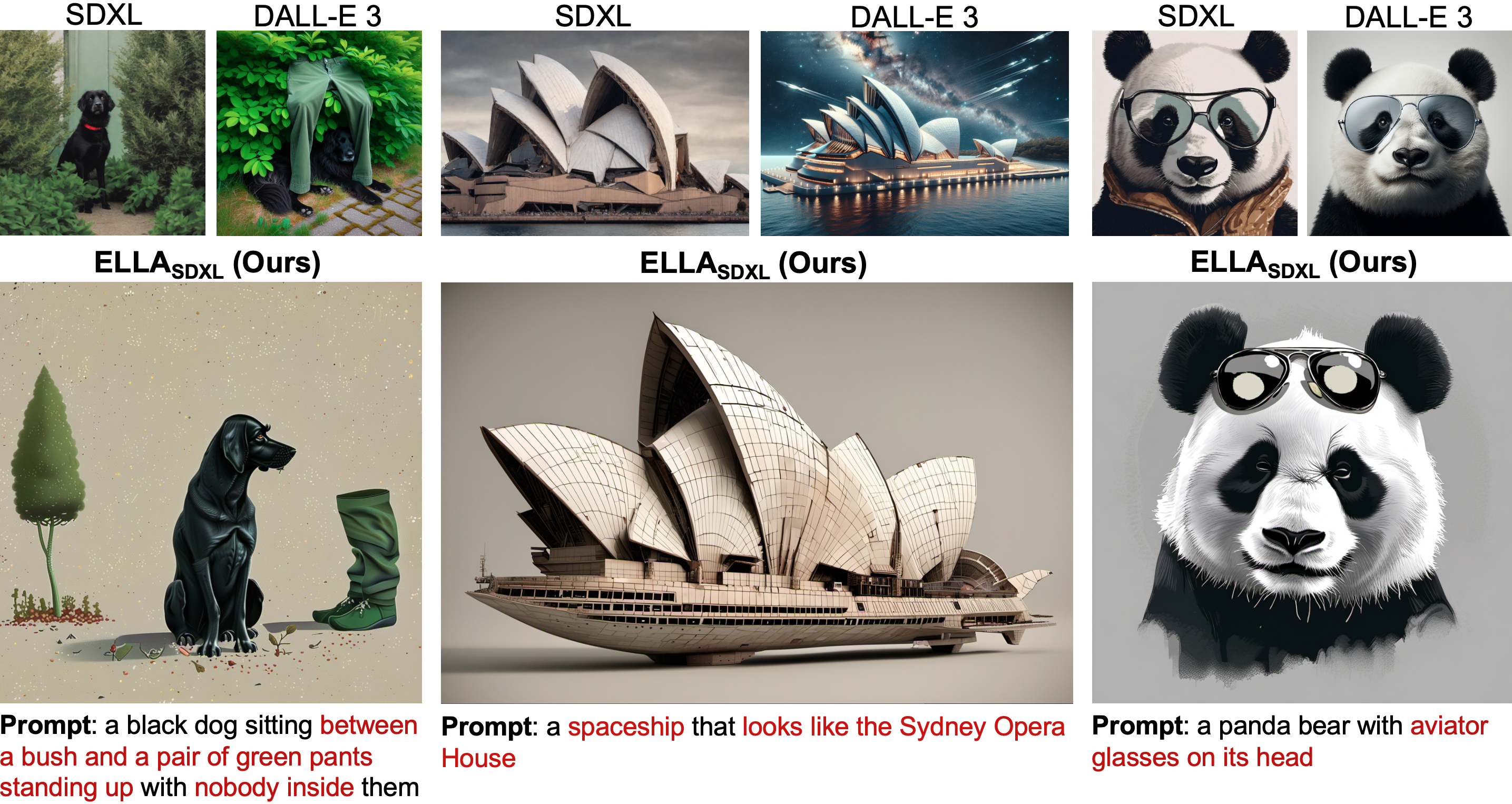}
  \vspace{1mm}
  \end{subfigure}
  
  \begin{subfigure}{1\linewidth}
    \includegraphics[width=0.99\linewidth]{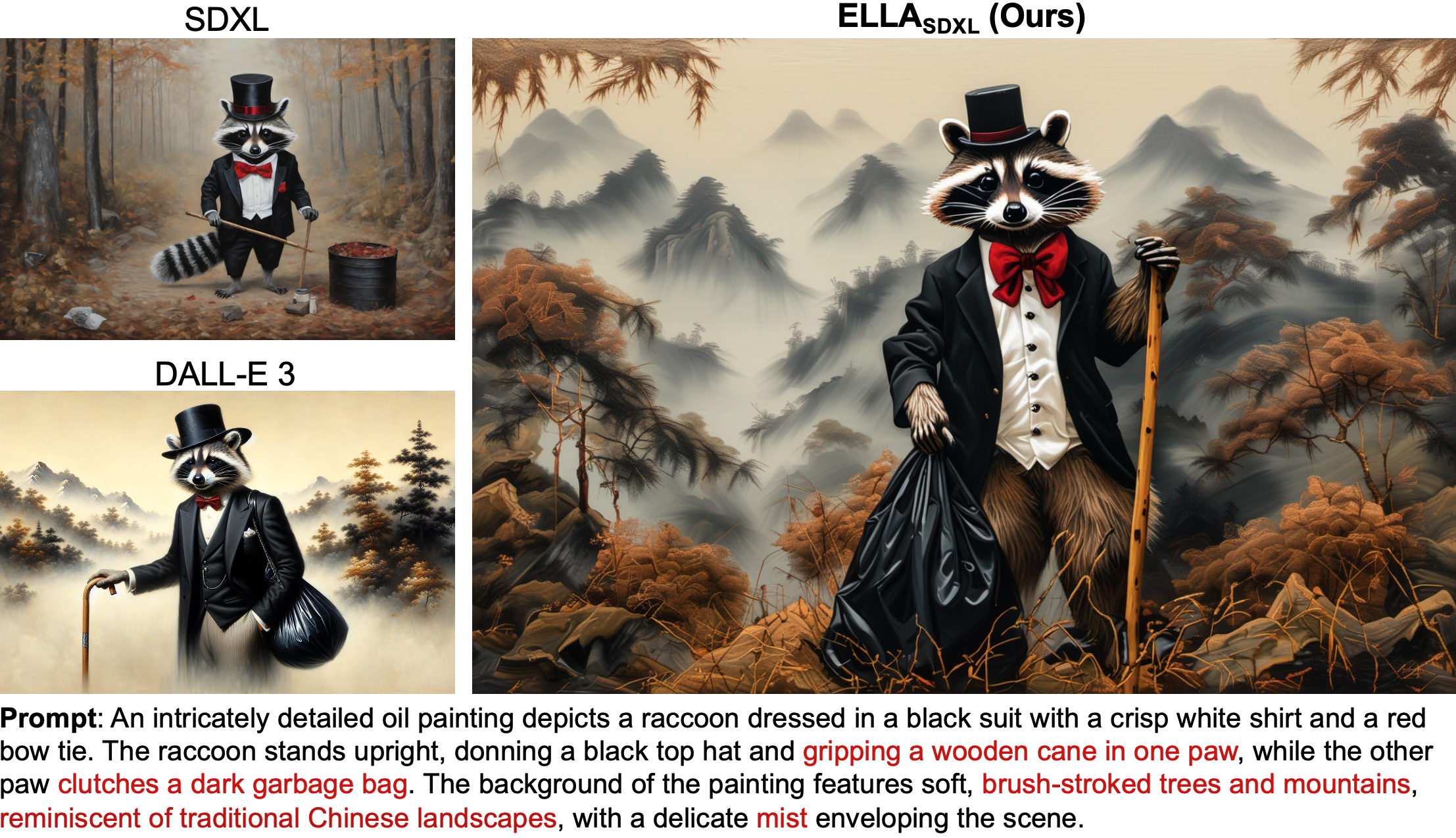}
  \end{subfigure}
  \caption{Comparison to SDXL~\cite{podell2023sdxl} and DALL-E 3~\cite{betker2023improving}. The prompts originate from PartiPrompts~\cite{yu2022scaling} (colored text denotes critical entities or attributes).}
  \vspace{-20pt}
  \label{fig:teaser}
\end{figure}

\begin{figure}[tb]
  \centering
  \includegraphics[width=1\linewidth]{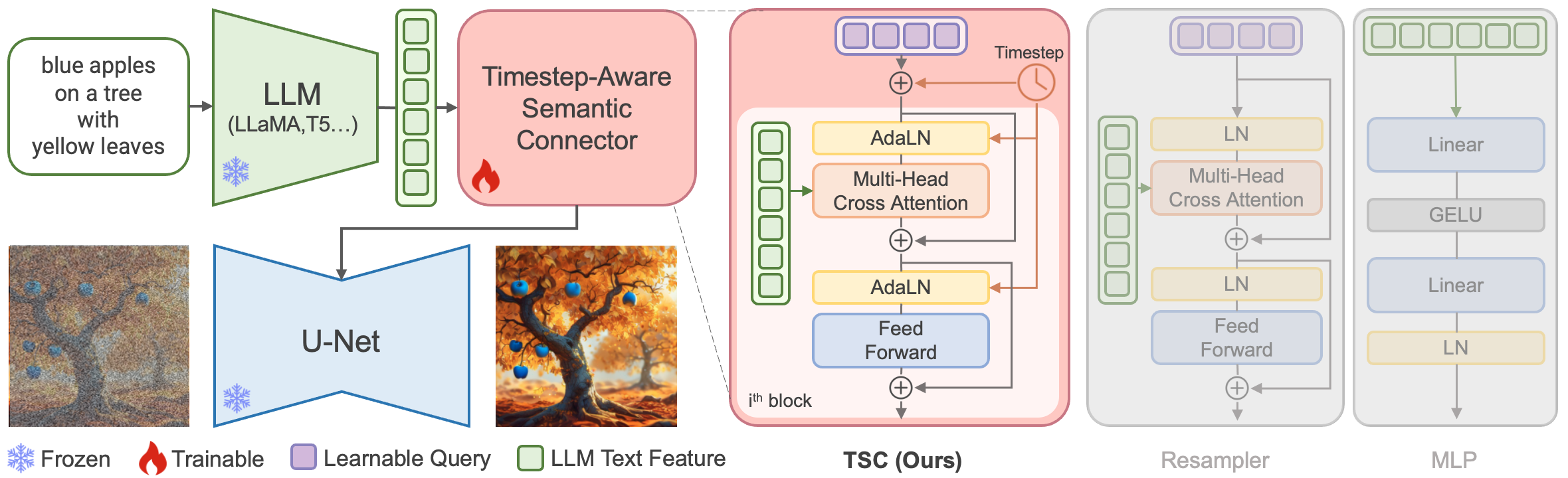}
  \caption{{\bf The overview of ELLA.} {\it Left:} The pipeline of our architecture. {\it Right:} The details of our TSC design alongside potential alternatives for connectors. We have conducted ablation studies on different connectors and finally selected TSC. }
  \vspace{-18pt}
  \label{fig:arch}
\end{figure}

\vspace{-20pt}
\section{Introduction}
\label{sec:intro}

In recent years, significant advancements have been made in text-to-image generation based on diffusion models.
These models~\cite{rombach2022high, podell2023sdxl, saharia2022photorealistic, ramesh2022hierarchical, betker2023improving} are capable of generating text-relevant images with high aesthetic quality, thereby driving the development of open-source community models and downstream tools. To adhere to prompt instructions, various existing models, especially~\cite{podell2023sdxl,rombach2022high,ramesh2022hierarchical}, employ the pre-trained CLIP~\cite{radford2021learning} model as a text encoder, which is trained on images with predominantly short text pairs. However, these models encounter difficulties in handling long dense prompts, particularly when the text describes multiple objects along with their distinct attributes and relationships. Some models~\cite{saharia2022photorealistic,wu2023paradiffusion} investigate the incorporation of powerful Large Language Models (LLM), such as T5~\cite{raffel2020exploring} and LLaMA-2~\cite{touvron2023llama}, with diffusion models to achieve a deeper level of language understanding in text-to-image generation. Imagen~\cite{saharia2022photorealistic} first demonstrates that text features from LLMs pre-trained on text-only corpora are remarkably effective in enhancing text alignment for text-to-image synthesis. Nonetheless, current models~\cite{wu2023paradiffusion, chen2023pixart, saharia2022photorealistic} that employ LLM as a text encoder necessitate the full training of U-Net~\cite{ronneberger2015u}, and ParaDiffusion~\cite{wu2023paradiffusion} even fine-tunes the pre-trained LLM. Aside from consuming vast computational resources, these models are difficult to integrate with the burgeoning community models and downstream tools~\cite{zhang2023adding,ye2023ip}.

To address the dense prompt understanding limitations of CLIP-based diffusion models, we propose a novel approach named ELLA, which incorporates powerful LLM in a lightweight and efficient manner. The architecture of ELLA is illustrated in~\cref{fig:arch}. Given pre-trained LLM and U-Net, we explore various semantic alignment connectors and train an adaptive connector, proposed as the Timestep-Aware Semantic Connector (TSC), on text-image pair data rich in information density. 
As observed by~\cite{balaji2022ediffi,hatamizadeh2023diffit,choi2022perception}, diffusion models typically predict low-frequency content during the initial stages of the denoising process and subsequently concentrate on high-frequency details towards the final stage. Consequently, we anticipate our alignment connector to initially extract text features at the low-frequency semantic level, which corresponds to the main objects and layout described in the prompt. Conversely, during the latter stages of denoising, we expect TSC to extract high-frequency semantics, which pertain to detailed attributes. 
The architectural design of our TSC is based on the resampler~\cite{alayrac2022flamingo}, and it instills temporal dependency by integrating the timestep in the Adaptive Layer Normalization~\cite{peebles2023scalable,perez2018film}. 
To facilitate TSC in dense information comprehension, we assemble highly informative text-image pair datasets, the captions of which are generated by state-of-the-art Multi-modal Language Language Models (MLLM). Once trained, TSC can seamlessly integrate community models and downstream tools such as LoRA~\cite{hu2021lora} and ControlNet~\cite{zhang2023adding}, improving their text-image alignment. 

Additionally, we introduce the Dense Prompt Graph Benchmark (DPG-Bench), a comprehensive dataset consisting of 1,065 lengthy, dense prompts, 
designed to assess the intricate semantic alignment capabilities of text-to-image models. In contrast to earlier benchmarks~\cite{Cho2024DSG, yu2022scaling, huang2024t2i}, DPG-Bench encompasses dense prompts that describe multiple objects, each characterized by a variety of attributes and relationships.
This benchmark also facilitates automatic evaluation using state-of-the-art MLLM. Extensive experiments on T2I-CompBench~\cite{huang2024t2i} and DPG-Bench demonstrate the superior semantic alignment of ELLA compared to existing SOTA T2I models~\cite{rombach2022high, podell2023sdxl, saharia2022photorealistic, ramesh2022hierarchical, betker2023improving}.

Our key contributions include:
\vspace{-5pt}
\begin{itemize}
    \item We propose a novel lightweight approach ELLA to equip existing CLIP-based diffusion models with powerful LLM.
    Without training of U-Net and LLM, ELLA improves prompt-following abilities and enables long dense text comprehension of text-to-image models.

    \item We design a Timestep-Aware Semantic Connector (TSC) to extract timestep-dependent conditions from the pre-trained LLM at various denoising stages. Our proposed TSC dynamically adapts semantics features over sampling time steps, which effectively conditions the frozen U-Net at distinct semantic levels.

    \item We introduce the DPG-Bench, comprising 1,065 lengthy, dense prompts characterized by a multitude of attributes and relationships.  
    Experimental results from user studies corroborate that the proposed evaluation metrics are highly correlated with human perception.
    
    \item Extensive results show that our ELLA exhibits superior performance in text alignment compared to existing state-of-the-art models, and significantly enhances the prompt-following capabilities of community models and downstream tools.
\end{itemize}

\section{Related Work}
\label{sec:related work}

\subsection{Text-to-Image Diffusion Models.}

Diffusion-based text-to-image models have exhibited remarkable enhancements in generating high-fidelity and diverse images.
These models need powerful text encoders to comprehend detailed image descriptions.
GLIDE\cite{nichol2021glide}, LDM\cite{rombach2022high}, DALL-E 2~\cite{ramesh2022hierarchical} and Stable Diffusion\cite{rombach2022high,podell2023sdxl} employ the pre-trained CLIP\cite{radford2021learning} model to extract text embeddings.
Imagen\cite{saharia2022photorealistic}, Pixart-$\alpha$\cite{chen2023pixart} and DALL-E 3\cite{betker2023improving} use pre-trained large language models (e.g., T5\cite{raffel2020exploring}) as text encoders, demonstrating that language text features exhibit a better understanding of text.
eDiff-I\cite{balaji2022ediffi} and EMU\cite{dai2023emu} use both CLIP and T5 embeddings as conditions. ParaDiffusion\cite{wu2023paradiffusion} proposes fine-tuning the LLaMA-2\cite{touvron2023llama} model during diffusion model training and using the fine-tuned LLM text features as the condition.
We equip the pre-trained CLIP-based models with LLM\cite{raffel2020exploring,touvron2023llama,zhang2024tinyllama} to enhance prompt following ability with the help of TSC.

\subsection{Compositional Text-to-Image Diffusion models.}

Various methods of compositional text-to-image diffusion models have been explored to better adhere to complex prompts.
Some works~\cite{liu2022compositional,bar2023multidiffusion,li2023divide,rassin2024linguistic,chefer2023attend,feng2023trainingfree,chen2024training,xie2023boxdiff,densediffusion} attempt to manipulate cross-attention maps or latents according to spatial or semantic constraints in the prompt. However, these methods are contingent upon the interpretability of the base models and can only achieve coarse and suboptimal control\cite{yang2024mastering,cao2023masactrl,hertz2022prompt,lian2023llm}.
Another potential solution\cite{huang2024t2i,xu2024imagereward,sun2023dreamsync,fang2023boosting} is to leverage image understanding feedback as reward to fine-tune the text-to-image models. These approaches are potentially constrained by the limitations in CLIP text understanding ability. Some studies\cite{zhong2023adapter,hao2024optimizing} employ LLMs to enhance the prompt or its corresponding embedding. 
\cite{cho2023visual,feng2023ranni,wang2024divide,yang2024mastering,lian2023llm,feng2024layoutgpt} harness the planning and reasoning ability of LLMs\cite{achiam2023gpt, touvron2023llama} to deconstruct prompts into multiple regional descriptions, serving as condition to guide the image generation process.
Our method enhances the base model's ability to follow prompts and can be seamlessly integrated with training-free methods\cite{feng2023trainingfree,chen2024training,yang2024mastering}.

\vspace{-10pt}
\section{Method}
\label{sec:method}
\subsection{Architecture Design}

To leverage the language understanding capability of LLM and the image generation potential of diffusion models and bridge them effectively, we design our ELLA as depicted in~\cref{fig:arch}. We consider the pre-trained LLM, such as T5~\cite{raffel2020exploring}, TinyLlama~\cite{zhang2024tinyllama} and LLaMA-2~\cite{touvron2023llama}, as the text encoder, which provides the comprehensive text feature to condition image generation. A carefully designed Timestep-Aware Semantic Connector (TSC) receives the text feature with arbitrary length as well as the timestep embedding, and outputs fixed-length semantic queries. These semantic queries are used to condition noisy latent prediction of the pre-trained U-Net through cross-attention. To improve the compatibility and minimize the training parameters, we leave both the text encoder of Large Language Models as well as the U-Net and VAE components frozen. The only trainable component is consequently our lightweight TSC module.

\noindent\textbf{Text Encoder.} ELLA is compatible with any state-of-the-art Large Language Models as text encoder, and we have conducted experiments with various LLMs, including T5-XL~\cite{raffel2020exploring}, TinyLlama~\cite{zhang2024tinyllama}, and LLaMA-2 13B~\cite{touvron2023llama}. The last hidden state of the language models is extracted as the comprehensive text feature. The text encoder is frozen during the training of ELLA. 
The detailed performance comparison of different Large Language Models is given in~\cref{subsec:ablation_study}.

\noindent\textbf{Timestep-Aware Semantic Connector (TSC).} This module interacts with the text features to facilitate improved semantic conditioning during the diffusion process. We investigate various network designs that influence the capability to effectively transfer semantic understanding.
\noindent\begin{itemize}
\vspace{-3pt}
  \item [-] 
  {\it MLP.} Following widely used converter design in LLaVA~\cite{liu2024visual}, we apply a similar MLP to map the text feature to image generation condition space. 
  \item [-]
  {\it Resampler.} As we connect two different modalities, MLP may not be sufficient enough to fuse and transfer the text feature. In addition, MLP is not a flexible design facing variant-length input and is difficult to scale up. We follow the Perceiver Resampler design from~\cite{alayrac2022flamingo} to learn a predefined number of latent input queries with transformer-based blocks. The latent input queries interact with frozen text features through cross-attention layers, which allows the module to tackle input text of arbitrary token length. 
  \item [-]
  {\it Resampler with timestep using Adaptive Layer Norm (AdaLN).} While keeping the text encoder and the diffusion model frozen, we expect our connector to provide highly informative conditions for noise prediction. Inspecting a given prompt for image generation, we notice that certain words describe the primary objects and corresponding attributes, while others may delineate details and image style. It is also observed that during image generation, the diffusion model initially predicts the main scene and subsequently refines the details~\cite{hatamizadeh2023diffit}. This observation inspires us to incorporate timestep into our resampler, allowing for the extraction of dynamic text features to better condition the noise prediction along with various diffusion stages. Our main results are based on this network design.
  \item [-]
  {\it Resampler with timestep using AdaLN-Zero.} AdaLN-Zero is another possible design to introduce timestep, which has been demonstrated more effective in DiT~\cite{peebles2023scalable}. However, this design is not the best choice in our framework and the experimental details can be found in~\cref{subsec:ablation_study}.
\end{itemize}

\subsection{Dataset Construction}
Most diffusion models adopt web datasets~\cite{schuhmann2021laion, kakaobrain2022coyo-700m} as the training dataset. However, the captions in these datasets, which are all alt-texts, often include overly brief or irrelevant sentences. This leads to a low degree of image-text correlation and a scarcity of dense semantic information. To generate long, highly descriptive captions, we apply the state-of-the-art MLLM CogVLM~\cite{wang2023cogvlm} as auto-captioner to synthesize image descriptions with the main objects and surroundings while specifying the corresponding color, texture, spatial relationships, {\it etc}. We show the vocabulary analysis~\cite{bird2009natural} of a subset of LAION~\cite{schuhmann2021laion} and COYO~\cite{kakaobrain2022coyo-700m} in~\cref{tab:info_dataset}, as well as our constructed CogVLM version. It can be observed that the alt-text in LAION/COYO contains significantly less information compared to the caption annotated by CogVLM. The latter features a greater number of nouns, adjectives, and prepositions, making it more descriptive in terms of multiple objects, diverse attributes, and complex relationships.
We filter images collected from LAION and COYO with an aesthetic score~\cite{improvedaestheticpredictor} over 6 and a minimum short edge resolution of 512 pixels. 
We annotated in total 30M image captions using CogVLM, which improves the matching between images and text and increases the information density of the captions.
To augment the diversity of prompt formats and improve the quality of generated images, we further incorporate 4M data from JourneyDB~\cite{sun2024journeydb} with their original captions. 

\begin{table}[h]
  \caption{{\bf Dataset information.} All the numbers in the table are the \textbf{average} results of each text in the dataset. Token statistics are calculated by CLIP tokenizer. The abbreviation is derived from NLTK. {\bf NN}: noun, including singulars, plurals and proper nouns. {\bf JJ}: adjective, including comparatives and superlatives. {\bf RB}: adverb, including comparatives and superlatives. {\bf IN}: preposition and subordinating conjunctions.
  }
  \label{tab:info_dataset}
  \centering
  {\scalebox{0.9}{\footnotesize
  \begin{tabular}{@{}lcccccc@{}}
    \toprule
    \textbf{Dataset} & \textbf{Words} & \textbf{NN} & \textbf{JJ/RB} & \textbf{IN} & \textbf{Tokens} \\
    \midrule
    LAION & 9.81 & 3.59 & 0.70 & 1.87 & 11.88 \\
    LAION-CogVLM & 49.87 & 15.51 & 8.06 & 6.26 & 62.33 \\
    \midrule
    COYO & 9.83 & 3.60 & 0.65 & 1.91 & 11.89 \\
    COYO-CogVLM & 50.71 & 15.71 & 8.06 & 6.38 & 63.05 \\
  \bottomrule
  \end{tabular}
  }}
\end{table}

\section{Benchmark}
\label{sec:benchmark}
\begin{figure}[b]
  \vspace{-10pt}
  \centering
  \begin{subfigure}{0.44\linewidth}
    \includegraphics[width=1\linewidth]{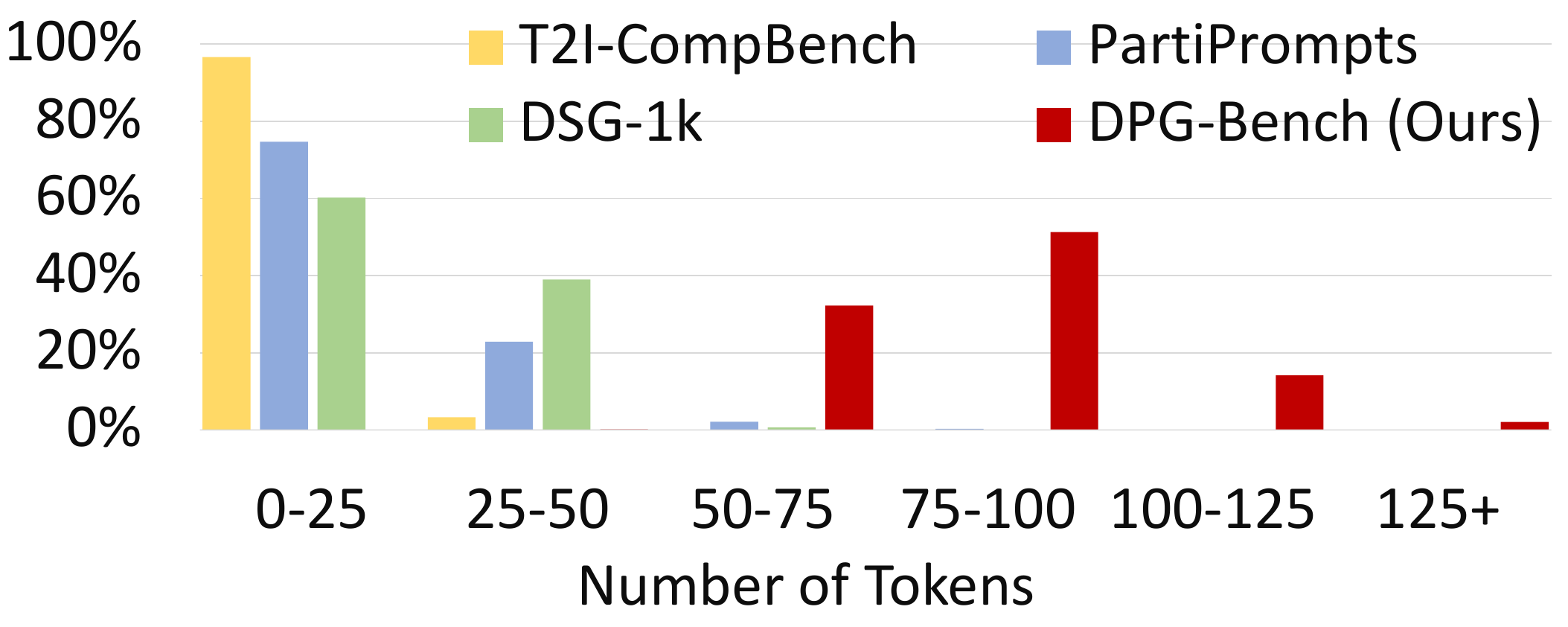}
  \end{subfigure}
  \begin{subfigure}{0.55\linewidth}
    \includegraphics[width=1\linewidth]{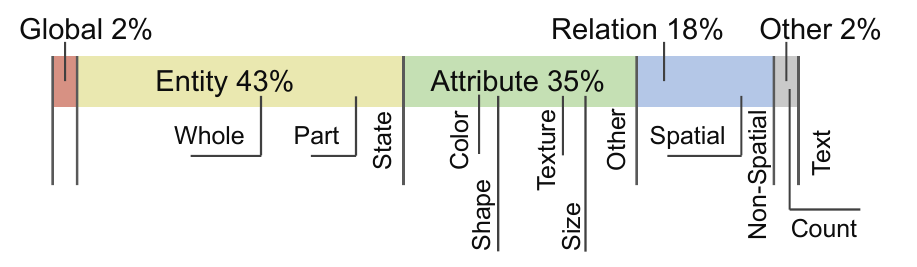}
  \end{subfigure}
  \caption{{\bf DPG-Bench Information. } {\it Left:} The token distribution of DPG-Bench and other benchmarks. {\it Right:} the level-1 categories and the level-2 categories.}
  \label{fig:dpg_bench}
\end{figure}

\vspace{-5pt}
Current benchmarks have not considered evaluating the generation models' ability to follow dense prompts. The average token length of prompts tokenized by CLIP tokenizer in previous benchmarks, such as T2I-CompBench~\cite{huang2024t2i} and PartiPrompts~\cite{yu2022scaling}, is about 10-20 tokens. Moreover, these benchmarks are not comprehensive enough to describe a diverse range of objects, as they contain limited words in each prompt and lack sufficient diversity of nouns throughout the entire benchmark, as listed in~\cref{tab:info_bench}. Therefore, we present a more comprehensive benchmark for dense prompts, called \textbf{Dense Prompt Graph Benchmark (DPG-Bench)}. Compared to previous benchmarks, as shown in~\cref{fig:dpg_bench}, DPG-Bench provides longer prompts containing more information.

\begin{table}[tb]
  \caption{{\bf Benchmark information.} {\bf \#DN} means the total distinct nouns in the benchmarks. All the numbers in the table are the {\bf average} results except for \#DN. Token statistics are calculated by CLIP tokenizer.}
  \label{tab:info_bench}
  \centering
  {\footnotesize
  \scalebox{0.9}{
  \begin{tabular}{@{}lccccccc@{}}
    \toprule
    \textbf{Benchmarks} & \textbf{\#DN} & \textbf{Words} & \textbf{NN}& \textbf{JJ/RB}& \textbf{IN}  & \textbf{Tokens} \\
    \midrule
    T2I-CompBench & 1447 & 9.60 & 3.40 & 1.36 & 0.87 & 12.65 \\
    PartiPrompts & 1421 & 9.11 & 3.45 & 0.94 & 1.36 & 12.20 \\
    DSG-1k & 2004 & 17.13 & 6.35 & 2.09 & 2.41 & 22.56 \\
    \midrule
    \textbf{DPG-Bench (ours)} & 4286 & 67.12 & 20.90 & 11.59 & 9.07 & 83.91 \\
  \bottomrule
  \end{tabular}
  }}
  \vspace{-10pt}
\end{table}

We gather source data from COCO~\cite{lin2014microsoft}, PartiPrompts~\cite{yu2022scaling}, DSG-1k~\cite{Cho2024DSG} and Object365~\cite{shao2019objects365}. For the data originating from the first three sources, we create long dense prompts based on the original short prompts. In the case of Object365, we randomly sampled 1-4 objects for each prompt according to its main category and subcategory, subsequently generating prompts with sampled objects. Given the source data, we instruct GPT-4 to specify the details of the scene, the attributes as well as the corresponding relationship of objects, creating long dense prompts with rich semantic information. All prompts are generated automatically by GPT-4 and verified by humans. With constructed dense prompts, we follow the pipeline of DSG, leveraging GPT-4 once more to generate corresponding tuple categories, questions, and graphs.
As shown in~\cref{fig:dpg_bench}, our benchmark provides a two-level category scheme, with 5 level-1 categories and 13 level-2 categories.

In conducting the assessment, each model is expected to generate 4 images for each given prompt. Subsequently, mPLUG-large~\cite{li2022mplug} is employed as the adjudicator to evaluate the generated images according to the designated questions. As for the score computation, our benchmark adheres to the principles of DSG. Ultimately, the mean score of a series of questions pertaining to a single prompt constitutes the prompt score, while the mean score of all prompt scores represents the DPG-Bench score.

\begin{figure}[t]
  \centering
  \includegraphics[width=1\linewidth]{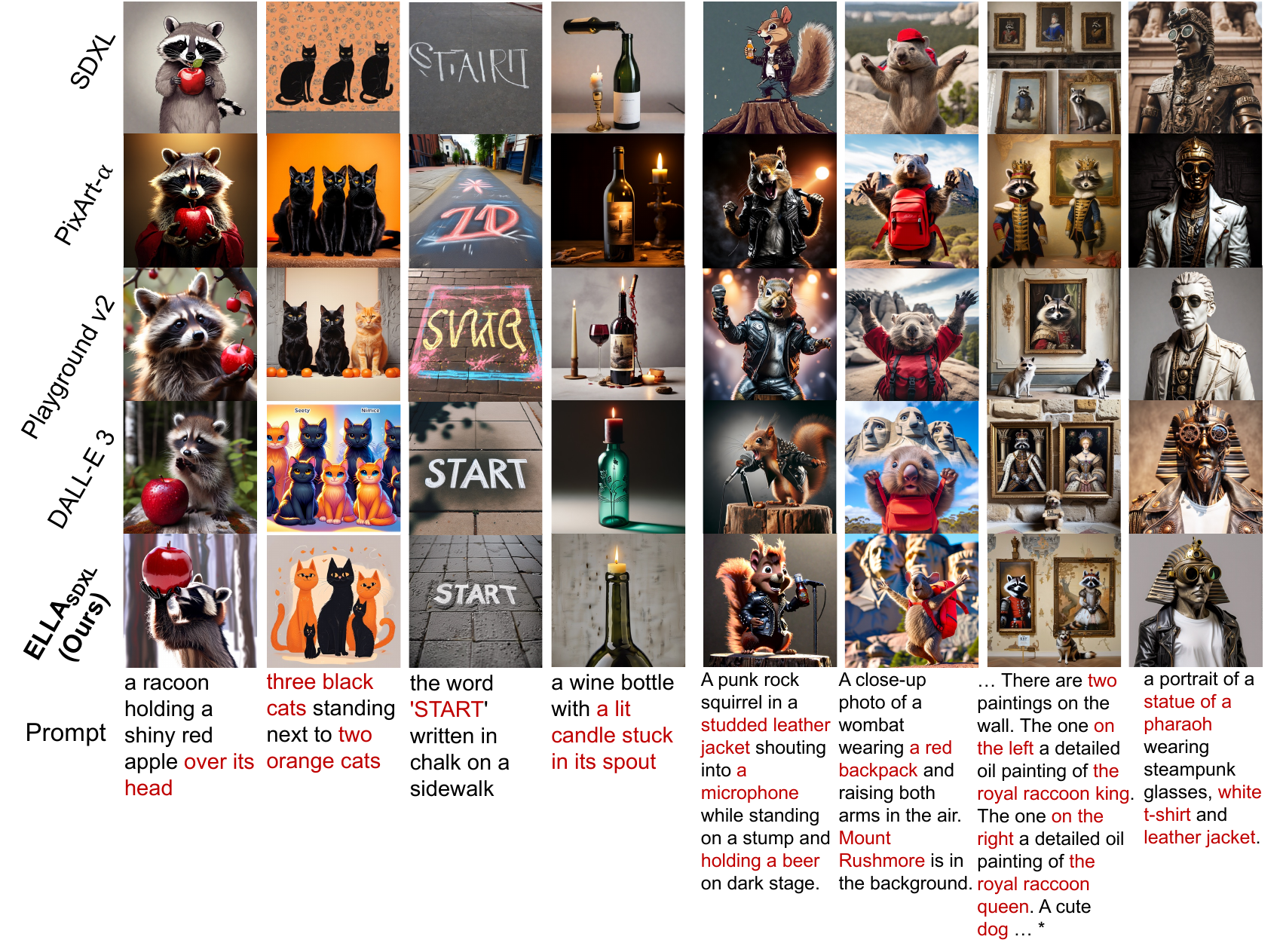}
  \vspace{-15pt}
  \caption{The comparison between ELLA, SDXL, PixArt-$\alpha$, Playground v2~\cite{playground-v2} and DALL-E 3. The left four columns only contain 1 or 2 entities, but the right four correspond to dense prompts with more than 2 entities. All prompts originate from PartiPrompts. The complete prompt with the mark * is written in the footnote.\protect\footnotemark[1]}
  \vspace{-10pt}
  \label{fig:exp_pic1}
\end{figure}
\footnotetext[1]{\scriptsize a wall in a royal castle. There are two paintings on the wall. The one on the left a detailed oil painting of the royal raccoon king. The one on the right a detailed oil painting of the royal raccoon queen. A cute dog looking at the two paintings, holding a sign saying 'plz conserve'}

\section{Experiments}
\label{sec:experiments}

\subsection{Implementation Details}
\textbf{Training Details.} We employ the encoder of T5-XL~\cite{raffel2020exploring}, a 1.2B model for text feature extraction. To be more compatible with models in the open-source community, we use SDv1.5~\cite{rombach2022high} and SDXL~\cite{podell2023sdxl} as base models for training. We set the training length of extracted text tokens as 128 to handle complex scene understanding in dense captions. Our models are trained on 34M image-text pairs with 512 resolution for text alignment. We further train our ELLA{\tiny\it{SDXL}} on 100K high-quality data with 1024 resolution for aesthetic improvements. The AdamW optimizer~\cite{loshchilov2017decoupled} is used with a weight decay of 0.01, a constant 1e-4 learning rate for ELLA{\tiny\it{SDv1.5}} and 1e-5 for ELLA{\tiny\it{SDXL}}. The final model is trained on 8 40G A100 for approximately 7 days for the ELLA{\tiny\it{SDv1.5}} and 14 days for ELLA{\tiny\it{SDXL}}. Our ELLA{\tiny\it{SDXL}} costs less than 80\% training time compared to PixArt-$\alpha$~\cite{chen2023pixart} (753 A100 GPU days). 

\begin{table}[h]
  \caption{Evaluation on T2I-CompBench with short compositional prompts. 
  The \textbf{bold} score denotes the best performance. Higher score for better performance.}
  \label{tab:t2i_1}
  \centering
  {\footnotesize
  \begin{tabular}{@{}lccccc@{}}
    \toprule
     \multirow{2}{*}{\textbf{Model}} & \multicolumn{3}{c}{\textbf{Attribute Binding}} & \multicolumn{2}{c}{\textbf{Object Relationship}} 
    \\
    \cmidrule(lr){2-4} \cmidrule(lr){5-6}
      & \textbf{Color} & \textbf{Shape} & \textbf{Texture} & \textbf{Spatial} & \textbf{Non-Spatial} \\
    \midrule
    SD v1.4 & 0.3765 & 0.3576 & 0.4156 & 0.1246 & 0.3079 
    \\
    
    SD v2 & 0.5065 & 0.4221 & 0.4922 & 0.1342 & 0.3096 
    \\
    Composable v2 & 0.4063 & 0.3299 & 0.3645 & 0.0800 & 0.2980
    \\
    Structured v2 & 0.4990 & 0.4218 & 0.4900 & 0.1386 & 0.3111
    \\
    Attn-Exct v2 & 0.6400 & 0.4517 & 0.5963 & 0.1455 & 0.3109
    \\
    GORS & 0.6603 & 0.4785 & 0.6287 & 0.1815 & \textbf{0.3193}
    \\
    DALL-E2 & 0.5750 & 0.5464 & 0.6374 & 0.1283 & 0.3043 
    \\
    PixArt-\it{$\alpha$} & 0.6886 & 0.5582 & \textbf{0.7044} & 0.2082 & 0.3179 
    \\
    \midrule
    SD v1.5 & 0.3750 & 0.3724 & 0.4159 & 0.1204 & 0.3088 
    \\
    \textbf{ELLA{\tiny\it{SDv1.5}}} & 0.6911 & 0.4938 & 0.6308 & 0.1867 & 0.3062 
    \\
    \midrule
    SDXL & 0.6369 & 0.5408 & 0.5637 & 0.2032 & 0.3110 
    \\
    \textbf{ELLA{\tiny\it{SDXL}}} & \textbf{0.7260} & \textbf{0.5634} & 0.6686 & \textbf{0.2214} & 0.3069
    \\
  \bottomrule
  \end{tabular}
  }
\end{table}
\label{subsec:performance_comparison}
\vspace{-30pt}

\subsection{Performance Comparison and Analysis}

\noindent\textbf{Alignment Assessment.} To evaluate our model 
on short compositional prompts, 
we first conduct experiments on a subset of T2I-CompBench~\cite{huang2024t2i} to assess the alignment between the generated images and text conditions in attribute binding and object relationships. We compare our models with recent Stable Diffusion~\cite{rombach2022high} v1.4, v1.5, v2 and XL~\cite{podell2023sdxl} model, Composable v2~\cite{liu2022compositional}, Structured v2~\cite{feng2022training}, Attn-Exct v2~\cite{chefer2023attend}, GORS~\cite{huang2024t2i}, PixArt-$\alpha$~\cite{chen2023pixart} and DALL-E 2~\cite{rombach2022high}. As shown in~\cref{tab:t2i_1}, our ELLA{\tiny\it{SDv1.5}} performance significantly surpasses its based model SD v1.5~\cite{rombach2022high} and is even comparable to SDXL in some subset of evaluation categories. The ELLA{\tiny\it{SDXL}} performs better than SDXL and the fully-tuned PixArt-$\alpha$ in most categories. It is observed that the MLLM caption model is highly sensitive to information such as color and texture in images. Therefore, such data greatly contributes to the learning of these attributes. 

\begin{table}[ht]
  \caption{{\bf Evaluation results on DPG-Bench.} Average score is the graph score based on the rule of DSG and the larger score is better. The other scores are the average of all questions in one category. The VQA answers are generated by mPLUG-large~\cite{li2022mplug}. {\bf \#Params} denotes trainable parameters. Higher score for better performance. The \textbf{bold} score for the best and the \textbf{\underline{bold underlined}} score for the second best.
  }
  \label{tab:DPG-Bench}
  \centering
  {\footnotesize
  \begin{tabular}{@{}l|c|c|ccccc@{}}
    \toprule
    \textbf{Model} & \textbf{\#Params} & \textbf{Average} & \textbf{Global} & \textbf{Entity} & \textbf{Attribute} & \textbf{Relation} & \textbf{Other} \\
    \hline
    
    SD v2 & 0.86B & 68.09 & 77.67 & 78.13 & 74.91 & 80.72 & 80.66 \\
    PixArt-\it{$\alpha$} & 0.61B & 71.11 & 74.97 & 79.32 & 78.60 & 82.57 & 76.96 \\
    Playground v2 & 2.61B & 74.54 & 83.61 & 79.91 & 82.67 & 80.62 & 81.22 \\
    DALL-E 3 & - & \textbf{83.50} & 90.97 & 89.61 & 88.39 & 90.58 & 89.83 \\
    
    \hline
    SD v1.5 & 0.86B & 63.18 & 74.63 & 74.23 & 75.39 & 73.49 & 67.81 \\
    \textbf{ELLA{\tiny\it{SDv1.5}}} & 0.07B & 74.91 & 84.03 & 84.61 & 83.48 & 84.03 & 80.79 \\
    \hline
    SDXL & 2.61B & 74.65 & 83.27 & 82.43 & 80.91 & 86.76 & 80.41 \\
    \textbf{ELLA{\tiny\it{SDXL}}} & 0.47B & \textbf{\underline{80.23}} & 85.90 & 85.34 & 86.67 & 86.16 & 87.41 \\
  \bottomrule
  \end{tabular}
  }
\end{table}

\begin{figure}[b]
  \vspace{-10pt}
  \centering
  \begin{subfigure}{0.49\linewidth}
    \includegraphics[width=1\linewidth]{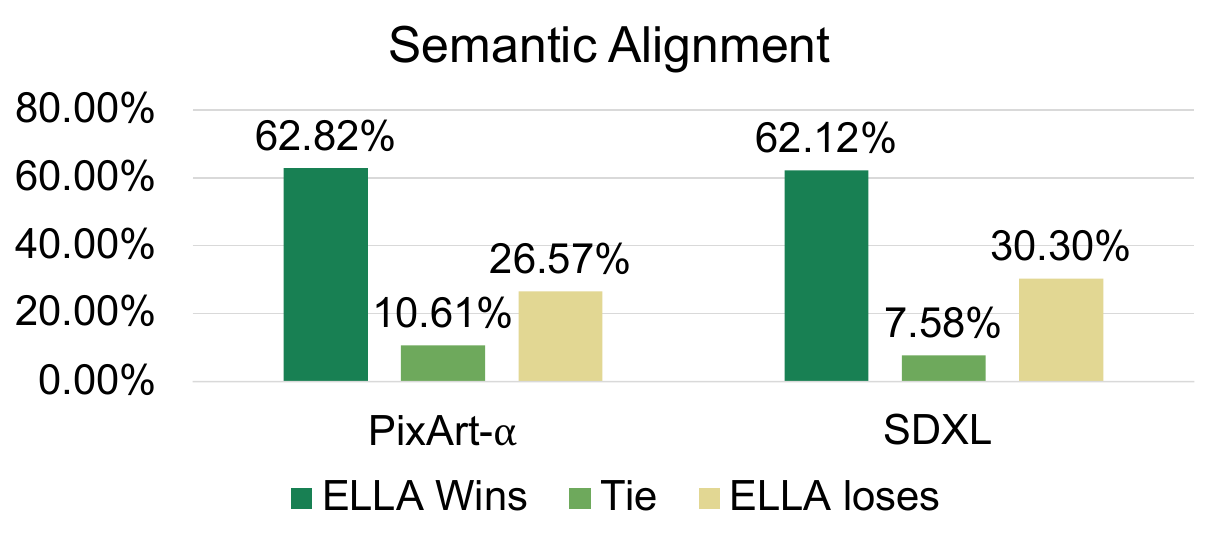}
    \label{fig:alignment}
  \end{subfigure}
  \begin{subfigure}{0.49\linewidth}
    \includegraphics[width=1\linewidth]{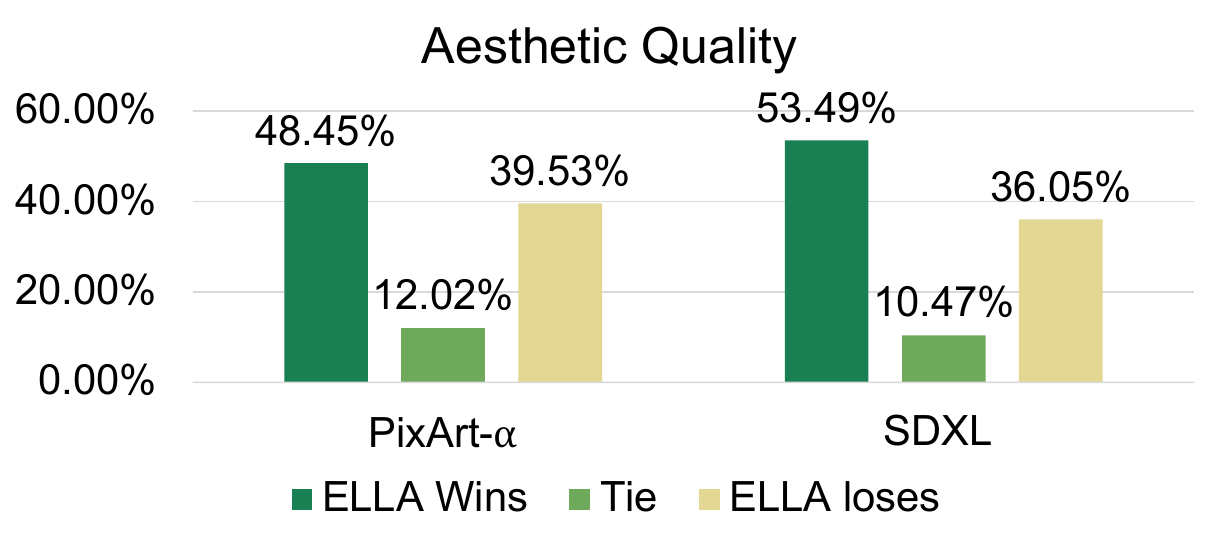}
    \label{fig:aesthetics}
  \end{subfigure}
  \vspace{-15pt}
  \caption{{\bf The results of user study.} The bar chart demonstrates that our model surpasses existing open-source models in terms of text-image alignment capabilities while maintaining a comparable aesthetic quality. }
  \vspace{-10pt}
  \label{fig:user_study}
\end{figure}

\begin{figure}[ht]
  \centering
  \includegraphics[width=1\linewidth]{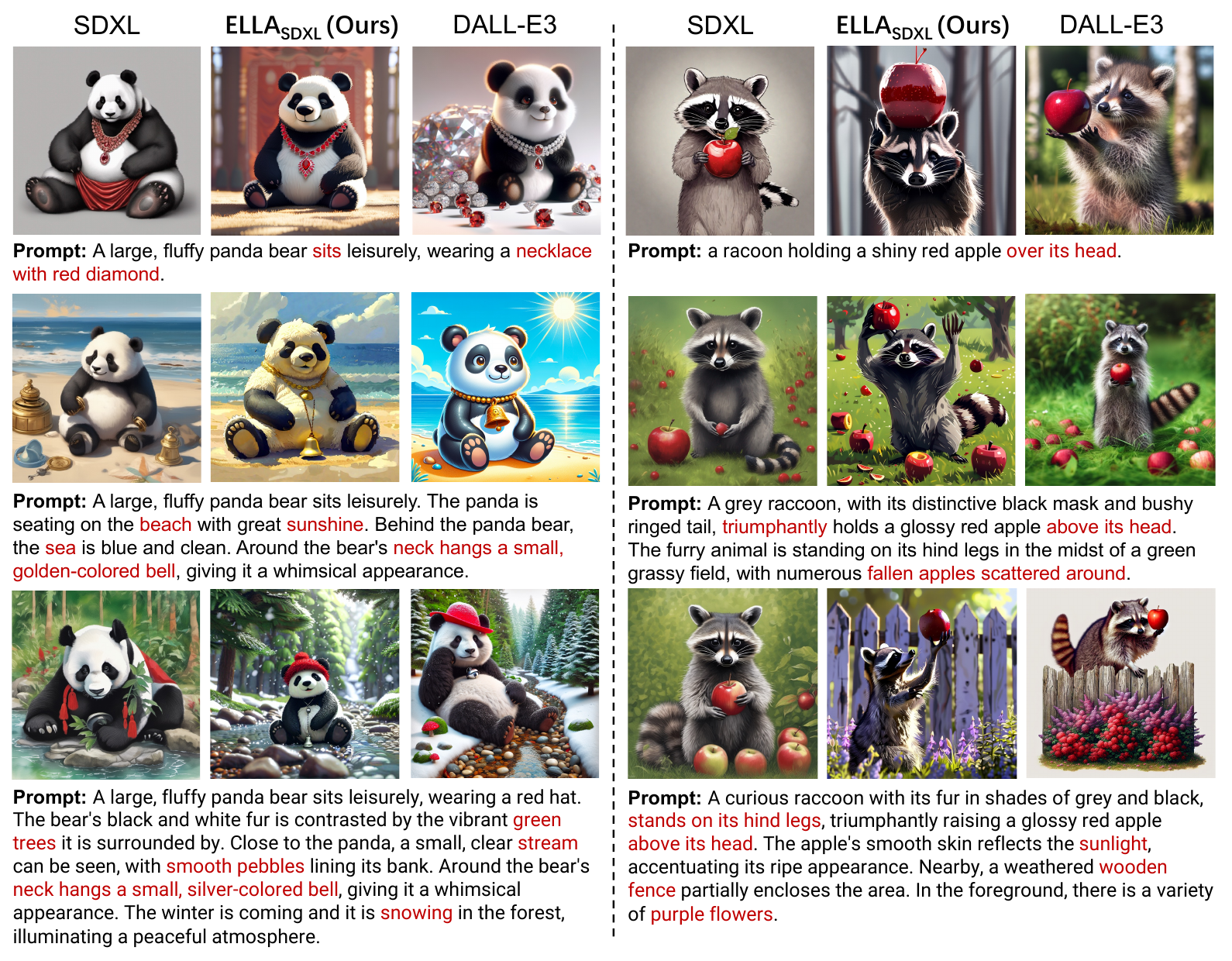}
  \caption{The comparison between SDXL, ELLA{\tiny\it{SDXL}}, and DALL-E 3 reveals their performance across varying levels of prompt complexity. Prompts range from simple to intricate from top to bottom. The results demonstrate that our model is capable of following both simple and complex prompts and generating fine-grained detail.}
  \vspace{-10pt}
  \label{fig:exp_pic2}
\end{figure}

\begin{figure}[tb]
  \centering
  \includegraphics[width=1\linewidth]{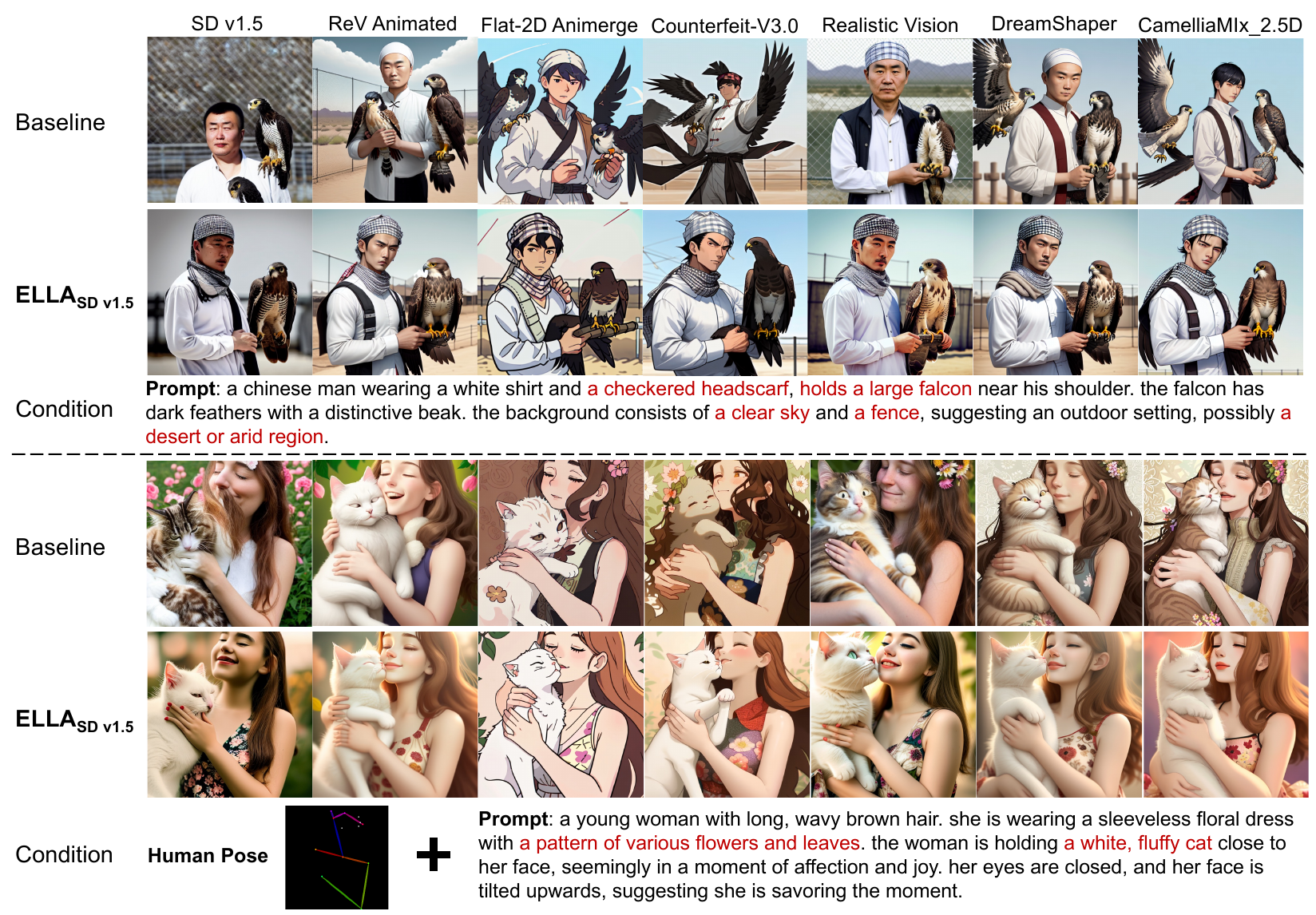}
  \caption{Qualitative results about ELLA{\tiny\it{SDv1.5}} with personalized models. We selected representative personalized models from CivitAI~\cite{civitai}, equipping them with ELLA{\tiny\it{SDv1.5}} to improve their prompt following ability.}
  \vspace{-15pt}
  \label{fig:community}
\end{figure}

\noindent\textbf{Dense Prompt Following.} 
To further measure a model's proficiency in following longer prompts laden with dense information, we test different models on our proposed DPG-Bench to evaluate semantic alignment for complex texts. As depicted in~\cref{tab:DPG-Bench}, our model has the smallest number of training parameters among previous models. Specifically, our ELLA{\tiny\it{SDv1.5}} and ELLA{\tiny\it{SDXL}} models require only 0.06B and 0.47B training parameters, respectively. Despite this, they demonstrate superior performance over other models across the evaluation metrics, with their performance only slightly trailing behind that of DALL-E 3. To better inspect the performance of different models, we illustrate the qualitative results on PartiPrompts in~\cref{fig:exp_pic1}. 
We also conduct experiments to incrementally increase the complexity of the prompt. As illustrated in~\cref{fig:exp_pic2}, both SDXL and DALL-E 3 exhibit difficulties in managing dense prompts, often conflating multiple objects and missing fine-grained details. In contrast, our ELLA model demonstrates a superior comprehension of the given prompt, generating images that capture more precise details.

\noindent\textbf{User Study.} While quantitative results can partially represent a model's performance, they may not be sufficiently comprehensive and accurate. Consequently, we conducted a user study to enrich the evaluation of the model. We employ image generation results of our DPG-Bench to examine the text-image alignment and aesthetic quality of CLIP-based SDXL, T5-based PixArt-$\alpha$ and our ELLA{\tiny\it{SDXL}}. For each prompt with 4 images generated by distinct models, we enlist 20 unique users to rank images based on semantic alignment and aesthetic quality. The user study results are reported in~\cref{fig:user_study}. We observe that human preferences align with our evaluation outcomes on DPG-Bench in~\cref{tab:DPG-Bench}, thereby attesting to the reliability of DPG-Bench. Across the board, our ELLA{\tiny\it{SDXL}} outperforms the current SOTA open-source models in terms of text-image alignment, while maintaining an aesthetic quality comparable to that of SDXL.

\noindent\textbf{Compatibility with Downstream Tools.}
As we freeze the original diffusion model in the training stage, ELLA can be seamlessly integrated into the downstream tools of Stable Diffusion. 
We equip six widely used community models from CivitAI~\cite{civitai} with ELLA to further improve their prompt following ability. As shown in~\cref{fig:community}, community models (e.g., LoRA~\cite{hu2021lora}, ControlNet~\cite{zhang2023adding}) can significantly benefit from ELLA, improving the semantic matching of generated images while maintaining a similar style to the original. 

\vspace{-10pt}
\subsection{Ablation Study}
\label{subsec:ablation_study}

\begin{table}[tb]
    \vspace{-10pt}
  \caption{Ablation study on LLM selection based on SD v1.5. CLIP represents the original SD v1.5. For LLM-based text encoder, resampler with 6 blocks is applied. }
  \label{tab:ablation_llm}
  \centering
  {\footnotesize
  \begin{tabular}{@{}lccccccc@{}}
    \toprule
    \multirow{2}{*}{\textbf{Text Encoder}} & 
    \multicolumn{3}{c}{\textbf{Attribute Binding}} & \multirow{2}{*}{\textbf{DPG-Bench$\uparrow$}} \\
    \cmidrule(lr){2-4}
      & Color$\uparrow$ & Shape$\uparrow$ & Texture$\uparrow$ & \\
     \hline
     \textbf{CLIP} & 
     0.3750 & 0.3576 & 0.4156 & 63.18 \\
     \hline
     \textbf{TinyLlama} & 
     0.4168 & 0.3922 & 0.4639 & 70.27 \\
     \textbf{LLaMA-2} &
     0.4468 & 0.3983 & 0.5137 & 72.05 \\
     \textbf{T5-XL} & 
     0.5570 & 0.4522 & 0.5195 & 71.70 \\
  \bottomrule
  \end{tabular}
  }
\end{table}

We perform ablation experiments to inspect the effect of different LLM selections and alternative architecture designs. All experiments in the ablation study are conducted based on SD v1.5. It is noted that with limited computational resources, we train for 140,000 optimization steps (about 1 epoch) in the ablation study and 280,000 steps in the main experiments.

\noindent\textbf{LLM Selection.} Considering that the structure and number of parameters of different Large Language Models may result in varying capabilities in semantic understanding, we conduct experiments with 1B T5-XL encoder, 1.1B TinyLlama, and 13B LLaMA-2 on the SD v1.5. We employ the same semantic alignment connector with 6 blocks of resampler. 
~\cref{tab:ablation_llm} reports the corresponding performance on a subset of T2I-CompBench~\cite{huang2024t2i} and DPG-Bench, which assess short and complex prompts understanding respectively. 
Our ELLA, when equipped with various Large Language Models, consistently demonstrates superior performance over SD v1.5, which is based on CLIP. We observe that for decoder-only models, LLaMA-2 with 13B parameters performs better than 1.1B TinyLlama on both short and complex prompts. Compared to decoder-only LLMs, the 1.2B T5-XL encoder shows significant advantages in short prompts interpretation while falling short of LLaMA-2 13B in comprehending complex text. We suspect that encoder models with bidirectional attention capabilities may capture richer text features, thereby providing more effective conditioning for image generation. However, the model scale of T5-XL may pose a constraint on its ability to understand intricate texts.

\begin{table}[tb]
  \caption{Ablation study on module design based on TinyLlama and SD v1.5. The last row represents our TSC design.
  }
  \vspace{-10pt}
  \label{tab:ablation_module}
  \centering
  {\footnotesize
  \scalebox{0.9}{
  \begin{tabular}{@{}lccccccccc@{}}
    \toprule
    \textbf{Module} & \multirow{2}{*}{\textbf{Norm}} & \textbf{Timestep} & \textbf{Trainable}  & \multicolumn{3}{c}{\textbf{Attribute Binding}} & \multirow{2}{*}{\textbf{DPG-Bench$\uparrow$}} \\
    \cline{5-7}
     \textbf{Arch.}  & & \textbf{Aware} & \textbf{Params}  & Color$\uparrow$ & Shape$\uparrow$ & Texture$\uparrow$ & \\
     \midrule
     MLP & LN & \ding{55} & 2.16M & 0.3262 & 0.3198 & 0.3957 & 62.55 \\
     \hline
      Resampler   & \multirow{2}{*}{LN}  & \multirow{2}{*}{\ding{55}} &\multirow{2}{*}{8.71M} & \multirow{2}{*}{0.3569} & \multirow{2}{*}{0.3343} & \multirow{2}{*}{0.4124} & \multirow{2}{*}{66.39} \\
     (1 block) &  &  &  &  &  &  \\
     \hline
      Resampler   & \multirow{2}{*}{LN}  & \multirow{2}{*}{\ding{55}} &\multirow{2}{*}{44.16M} & \multirow{2}{*}{0.4168} & \multirow{2}{*}{0.3922} & \multirow{2}{*}{0.4639} & \multirow{2}{*}{70.27} \\
     (6 blocks) &  &  &  &  &  &  \\
     \hline
     Resampler   & \multirow{2}{*}{AdaLN-Zero}  & \multirow{2}{*}{\ding{51}} &\multirow{2}{*}{73.91M} & \multirow{2}{*}{0.4774} & \multirow{2}{*}{0.3810} & \multirow{2}{*}{0.4964} & \multirow{2}{*}{70.43} \\
     (6 blocks) &  &  &  &  &  &  \\
     \hline
     Resampler & \multirow{2}{*}{AdaLN}  & \multirow{2}{*}{\ding{51}} & \multirow{2}{*}{66.82M} & \multirow{2}{*}{0.5014} & \multirow{2}{*}{0.4253} & \multirow{2}{*}{0.5175} & \multirow{2}{*}{72.91} \\
     (6 blocks) &  &  &  &  &  &  \\
  \bottomrule
  \vspace{-30pt}
  \end{tabular}
  }}
  
\end{table}

\noindent\textbf{Module Network Design.} We explore various network designs for our ELLA module to examine the effectiveness of our chosen network architecture. To this end, we conduct experiments on MLP, resampler, and resampler with timestep using AdaLN-Zero and AdaLN.
All experiments are conducted with TinyLlama. ~\cref{tab:ablation_module} compares the performance of different network designs. It is observed that at a similar model scale, the transformer-based module is more effective in transferring the capabilities of language models to diffusion models than MLP. In addition, transformer blocks are more flexible for scaling up than MLP, which facilitates the upscaling of the module. 
We also conduct a comparative analysis of the scaled-up resampler with 6 blocks, incorporating timestep with both AdaLN-Zero and AdaLN. The latter configuration represents our final design for TSC, which performs the best on evaluation metrics. Although with more trainable parameters, experimental results illustrate that AdaLN-Zero underperforms AdaLN in our situation. 
AdaLN-Zero initializes each resampler block as the identity function, potentially weakening the contribution of LLM features to the final condition features. 

\begin{figure}[tb]
  \centering
  \includegraphics[width=1\linewidth]{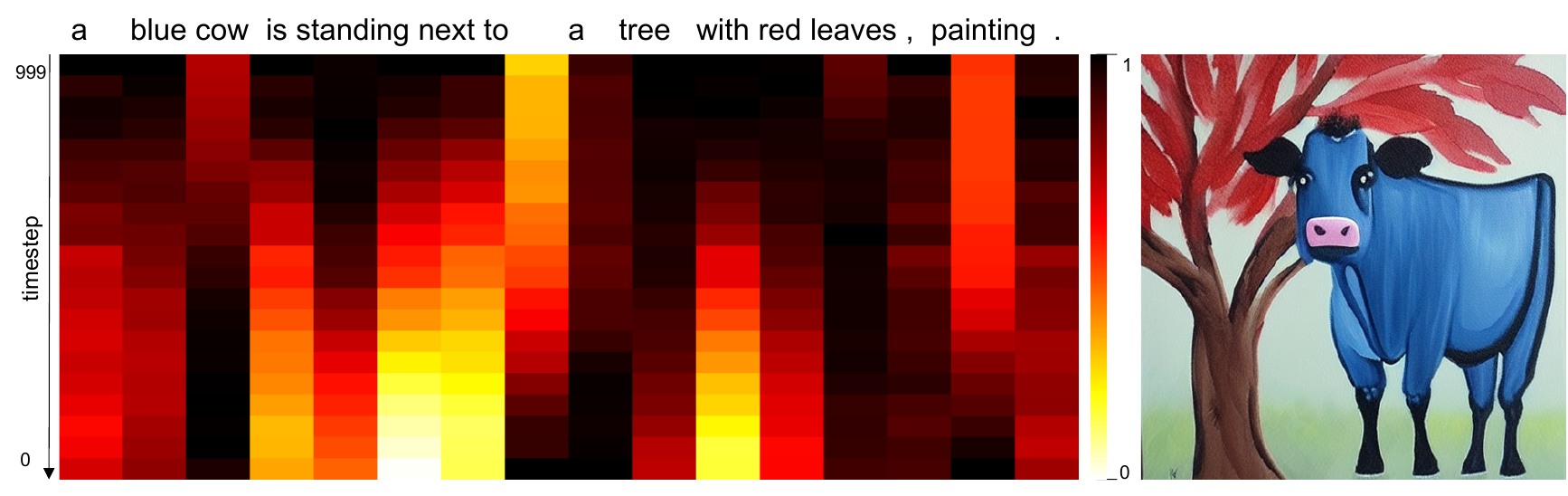}
  \vspace{-10pt}
  \caption{Visualization of the relative variation in attention scores between the text tokens feature and learnable queries, as influenced by the timestep.}
  \label{fig:exp_analysis}
  \vspace{-15pt}
\end{figure}

To analyze the extracted timestep-dependant semantic features of TSC, we visualize the relative variation in attention scores between text tokens and learnable queries at different denoising timesteps in~\cref{fig:exp_analysis}.
Across the diffusion timestep, for each text token in the prompt, we calculate the attention and normalize it by the maximum attention score of all timesteps.
Each column corresponds to a single token, demonstrating the temporal evolution of the significance attributed to each token.
It is observed that the attention values of words corresponding to the primary color (i.e., blue and red) and layout (i.e., standing next to) of the image are more pronounced at higher noise levels, during which the fundamental image formation transpires. 
In contrast, at lower noise levels, diffusion models generally predict high-frequency content, wherein the attention values of pertinent words (i.e., painting) that describe the image style become more prominent. 
This observation attests to the effectiveness of our proposed TSC in extracting semantic features from LLM across sampling timesteps. 
Furthermore, we note that the attention scores of the main entities, such as the cow and the tree, remain consistently strong throughout the denoising process, suggesting that diffusion models constantly focus on the primary entities during image construction.

\vspace{-10pt}
\section{Conclusion and Limitation}
\label{sec:conclusion-limitation}
This paper introduces ELLA, a method that equips current text-to-image diffusion models with state-of-the-art Large Language Models without the training of LLM and U-Net. We design a lightweight and adaptive Timestep-Aware Semantic Connector (TSC) to effectively condition the image generation process with comprehensive prompts understanding from LLM. With our ELLA, the diffusion model can generate high-fidelity and accurate images according to the given long prompts with dense information. In addition, we introduce a benchmark, rewritten from various data sources, to better evaluate the model performance on long dense prompts automatically. Experimental results demonstrate that ELLA outperforms current state-of-the-art text-to-image models in prompt following ability and can be easily incorporated with community models and downstream tools. With enhanced text-image alignment, our approach sheds light on image editing in future work. Furthermore, we also plan to investigate the integration of MLLM with diffusion models, enabling the utilization of interleaved image-text input as a conditional component in the image generation process. 
In terms of limitations, our training captions are synthesized by MLLM, which are sensitive to the entity, color, and texture, but are usually unreliable to the shape and the spatial relationship. On the other hand, the aesthetic quality upper bound of generated images may be limited by the frozen U-Net in our approach. These can be further addressed in the future.

\section*{Acknowledgements}
\label{sec:acknowledgement}

We thank Zebiao Huang, Huazheng Qin, Lei Huang, Zhengnan Lu, and Chen Fu for their support on data construction. We thank Fangfang Cao and Gongfan Cheng for their input on evaluation. We thank Keyi Shen and Dongcheng Xu for their support in accelerating training.


\clearpage  

%
%
\bibliographystyle{splncs04}
\bibliography{main}
\end{document}